\documentclass[letterpaper, 10 pt, conference]{ieeeconf}  % Comment this line out
\IEEEoverridecommandlockouts                              % This command is only
\usepackage{graphics}
\usepackage{epsfig} % for postscript graphics files
\usepackage{cite}
\usepackage{hyperref}

\title{\LARGE \bf
Fire SSD: Wide Fire Modules based Single Shot Detector on Edge Device
}

\author{HengFui, Liau,$^{1}$, Nimmagadda, Yamini,$^{2}$  and YengLiong, Wong$^{1}$\\
\textit{\{heng.hui.liau,yamini.nimmagadda,yeng.liong.wong\}@intel.com,}
\thanks{$^{1}$ Hengfui Liau and YengLiong Wong are with Intel's Internet of Things Group, Bayan Lepas, Penang, Malaysia.}
\thanks{$^{2}$Nimmagadda Yamini is with Intel's Software and Service Group, Hillsboro, Oregon, United States.}
}

\begin{document}

\maketitle
\thispagestyle{empty}
\pagestyle{empty}

%%%%%%%%%%%%%%%%%%%%%%%%%%%%%%%%%%%%%%%%%%%%%%%%%%%%%%%%%%%%%%%%%%%%%%%%%%%%%%%%
\begin{abstract}

With the emergence of edge computing, there is an increasing need for running convolutional neural network based object detection on small form factor edge computing devices with limited compute and thermal budget for applications such as video surveillance. To address this problem, efficient object detection frameworks such as YOLO and SSD were proposed. However, SSD based object detection that uses VGG16 as backend network is insufficient to achieve real time speed on edge devices. To further improve the detection speed, the backend network is replaced by more efficient networks such as SqueezeNet. Although the speed is greatly improved, it comes with a price of lower accuracy. In this paper, we propose an efficient SSD named “Fire SSD”. We augment the SqueezeNet based SSD framework with residual connection and group convolution. We design three new network modules to improve the accuracy of SqueezeNet based SSD while maintaining the real-time speed performance on CPU. The Fire SSD achieves 70.5mAP on Pascal VOC 2007 test set. Fire SSD achieves the speed of 31.7FPS on low power mainstream CPU and is about 6 times faster than SSD300 and has about 4 times smaller model size. Fire SSD also achieves 39.8FPS on integrated GPU.

\end{abstract}

%%%%%%%%%%%%%%%%%%%%%%%%%%%%%%%%%%%%%%%%%%%%%%%%%%%%%%%%%%%%%%%%%%%%%%%%%%%%%%%%
\section{INTRODUCTION}

Object detection is one of the main areas of research in computer vision and has lot of real word applications.  In recent years, object detection is widely used in several fields including but not limited to digital surveillance, autonomous vehicles, smart city, industrial automation, and smart home. With increasing need for protecting privacy and security, and for applications with limited network connectivity, edge computing devices (referred as ‘edge’ throughout this paper) are being deployed for local processing and inference of visual data. Edge devices are often thin clients and internet of things (IoT) systems with limited power and memory budget. To perform object detection on edge devices, there is a need for real-time inference speed, small model size and high energy efficiency.  To be economical and widely deployable, this object detector must operate on low power CPUs, integrated GPUs, or accelerators such as VPU~\cite{myriad} that dissipate far less power than powerful discrete GPUs used for benchmarking in typical computer vision experiments.

To address the need for running CNN based object detection in real-time, various object detection methods have been proposed~\cite{DPM}~\cite{rcnn}~\cite{rcnns}~\cite{fast}~\cite{FRCNN}~\cite{SSD}~\cite{YOLO}~\cite{YOLO2}. Among various object detection methods, Single Shot Multibox Detector (SSD)~\cite{SSD} has received a lot of attention because it is relatively fast and has an interchangeable backend network.  SSD is robust to scale variations because it makes use of multiple convolution layers at different scales for object detection.  Although the SSD300 and SSD512 perform well in detection accuracy, they still cannot fit within the constraints of edge devices. As an example, the model size of SSD300 that uses VGG16 network~\cite{DCN} as backend is 103MB. This makes the network model less efficient for distribution across edge devices over the network. Various SSD variants~\cite{dssd}~\cite{dsob}~\cite{rain}~\cite{fssd} have been proposed but most of the works focused on accuracy instead of efficiency. In complex visual use-cases, tasks such as video decode and image pre-processing are run on CPU or integrated GPU of edge devices and deep learning is offloaded to add-on accelerators such as FPGAs. To qualify as edge devices, low-power FPGAs with less than 10MB on-chip memory are used for such deployments~\cite{a10}. The large network size also limits the object detector being deployed on these accelerators. Second is the throughput of the object detector. SSD300 achieves 46FPS on Titian X~\cite{SSD}. However, edge devices may not have the luxury of using a 250W GPU because they are typically smaller in size and have very low power and thermal envelop. 

To further improve the throughput of SSD, the backend network which acts like a feature extractor is replaced by more efficient network architectures, such as SqueezeNet~\cite{SqueezeNet}, MobileNet~\cite{MobileNets}~\cite{xc}, ShuffleNet~\cite{Shufflenet} etc.  Iandola et al.~\cite{SqueezeNet} proposed SqueezeNet to address the issue of high parameter count for conventional neural networks. SqueezeNet reduces the number of parameter by replacing the conventional $3\times3$ convolutional layer by Fire Module. Fire module reduces the number of channels of the input data by employing $1\times1$ convolutional layer before applying a $3\times3$ convolutional and $1\times1$ convolutional layers in parallel with $\frac{1}{2}$ of the targeted number of channels. MobileNet proposes depth-wise separable convolution to reduce the number of computation operations significantly. ShuffleNet proposes pointwise group convolutions and channel shuffle to greatly  reduce the computations while maintaining accuracy~\cite{Shufflenet}. Replacing the backend network with SqueezeNet or MobileNet reduces the computation time significantly but it comes at a price of accuracy degradation. In this work, we propose an efficient SSD named Fire SSD that is both fast and accurate. The key features of Fire SSD are as follows:  
\begin{itemize}
\item \textbf{Wide fire module}(WFM)inspired by SqueezeNet, Inception~\cite{inception}, ResNext~\cite{ART} and ShuffleNet. Group convolution~\cite{ART}~\cite{ImageNet}~\cite{xc} has proven not only to reduce the computation complexity, but also improves the accuracy at the same time.  We improved the original Fire module by replacing the $3\times3$ and $1\times1$ expand convolution layer with group convolution layer. We studied the performance difference between group $1\times1$ convolution and $1\times1$ convolution and the experimental result shows that group number for $1\times1$ group convolution need to be small to maintain good accuracy.   
\item \textbf{Dynamic residual mbox detection layer}. In SSD framework, six feature maps with different resolutions are extracted from the backend network. Each feature map is processed by a single $3\times3$ convolutional layer for bounding box regression and object class prediction.  Feature maps at higher layers are responsible for detecting small scale objects. SSD does not perform well on detecting small scale objects because the corresponding feature maps are not deep enough and extracted features are not semantic enough to detect small scale objects. VGG16 and ResNet~\cite{DRL} show that deeper features improve the classification accuracy.  We stacked different number of convolution layers for feature maps at different levels.  To keep the model size and computational complexity small, we use Wide Fire Module (WFM). To compensate the gradient vanishing effect as the number of convolutional layers grow, a residual connection is introduced to connect upper and lower WFM modules.
\item \textbf{Normalized and dropout module}. The Normalized and Dropout Module (NDM) serves two purposes. First, NDM normalizes the gradient from different levels of feature maps by using batch normalization layer~\cite{BN}. This effectively eliminates the need of implementing L2 normalization layer for conv4\_3 feature map as described in SSD. Secondly, the dropout layer ~\cite{ImageNet}~\cite{Dropout}~\cite{drop2} helps regularize the training and performs data augmentation at different feature maps. Experimental result shows that NDM is a powerful generalization method and improves the accuracy by  1.4\%.
\end{itemize}

We evaluate our models on PASCAL VOC dataset. Fire SSD achieves 70.5mAP on Pascal VOC 2007 test set ~\cite{PASCALVOC}. Fire SSD achieves the speed of 31.7FPS on low power Intel Core\texttrademark CPU which is about 6 times faster than SSD300. We also implemented out network in Intel\textsuperscript{\tiny\textregistered} OpenVINO\texttrademark ~\cite{OpenVINO} and measured performance on a small form factor PC, Intel\textsuperscript{\tiny\textregistered} Skull Canyon ~\cite{Skull}. The result shows that Fire SSD achieved real-time speed on CPU and 39.8FPS on integrated GPU which is available in most of the personal computers. As shown in Figure 1, Fire SSD aims to hit the sweet spot between the CPU speed and accuracy which is the second quadrant of Figure 1.  

\section{RELATED WORK}

\subsection{Convolutional Neural Network for Object Detection}

SSD, YOLO~\cite{YOLO} and YOLOv2~\cite{YOLO2} are among the most popular fast object detection approaches. YOLO uses a single feedforward convolutional network to directly predict the object classes and location. Redmon et al.~\cite{YOLO2} proposed YOLOv2 to improve YOLO in several aspects. Redmon et al. also proposed a YOLOv2 with small model size named tiny YOLOv2. Tiny YOLOv2 only has 41\% of the number of parameters compared to YOLOv2 with 12mAP lower accuracy. Liu et al.~\cite{SSD} propose the SSD method, which spreads out the anchors of different scales to multiple layers within the backend network and uses feature maps from different layers to focus on predicting objects of a certain scale. SSD detects objects directly from different backend networks’ feature maps, it can achieve real-time object detection on high end GPUs and processes faster than most of other state-of-the-art object detectors.

\subsection{SqueezeNet}

Iandola et al.~\cite{SqueezeNet} proposed SqueezeNet to address the issue of high parameter count of conventional neural network. The SqueezeNet model has only 4.8MB of parameters and it matches AlexNet accuracy on ImageNet. The SqueezeNet model comprises of Fire Module. In a Fire Module, the input tensor of size $H \times W \times C$ is first downsampled using 1x1 convolution filter to reduce the channel size to $\frac{C}{4}$. Next, a $3\times3$ convolution filter with $\frac{C}{2}$ channels together with a parallel $1\times1$ convolutional filter with $\frac{C}{2}$ channels is used to fuse spatial information.  Given an H x W x C input and output size, a $3\times3$ convolutional layer requires $9 \times C^2$ parameters and $9 \times H \times W \times C^2$ computations, while Fire Module requires only $\frac{3}{2} \times C^2$ parameters and $\frac{3}{2} \times H \times W \times C^2$ computations.

\section{Network Architecture}

Fire SSD is based on SSD network. The backend network is replaced by residual SqueezeNet~\cite{SqueezeNet}. After Fire 8 layers, 10 WFM modules are appended along with a one $3\times3$ convolutional layer. Our purposed architecture is shown in Figure 2. Fire SSD consists of a backend network and six Multibox feature branches. The appended WFM modules are summarized in Table 1.

\begin{table}[ht]
\caption{Fire SSD: Appended wide fire modules and pooling layers}
\label{table_1}
\begin{center}
\begin{tabular}{|c||c||c||c|}
\hline
\textbf{Layers} & \textbf{Output Size} & \textbf{Stride} & \textbf{Channel}\\
\hline
Input (From Fire8) & $38 \times 38$ & - & 512\\
\hline
Pool8 & $19 \times 19$ & 2 & 512\\
\hline
Fire9 & $19 \times 19$ & 1 & 512\\
\hline
Fire10 & $19 \times 19$ & 1 & 512\\
\hline
Pool10 & $10 \times 10$ & 2 & 512\\
\hline
Fire11 & $10 \times 10$ & 1 & 512\\
\hline
Fire12 & $10 \times 10$ & 1 & 512\\
\hline
Pool12 & $5 \times 5$ & 2 & 512\\
\hline
Fire13 & $5 \times 5$ & 1 & 512\\
\hline
Fire14 & $5 \times 5$ & 1 & 512\\
\hline
Fire15 & $3 \times 3$ & 2 & 512\\
\hline
Fire16 & $3 \times 3$ & 1 & 512\\
\hline
Conv17 & $1 \times 1$ & 1 & 512\\
\hline
\end{tabular}
\end{center}
\end{table}

\begin{figure*}[ht]
%\begin{center}
\centering
\begin{center}
\includegraphics[width=\textwidth]{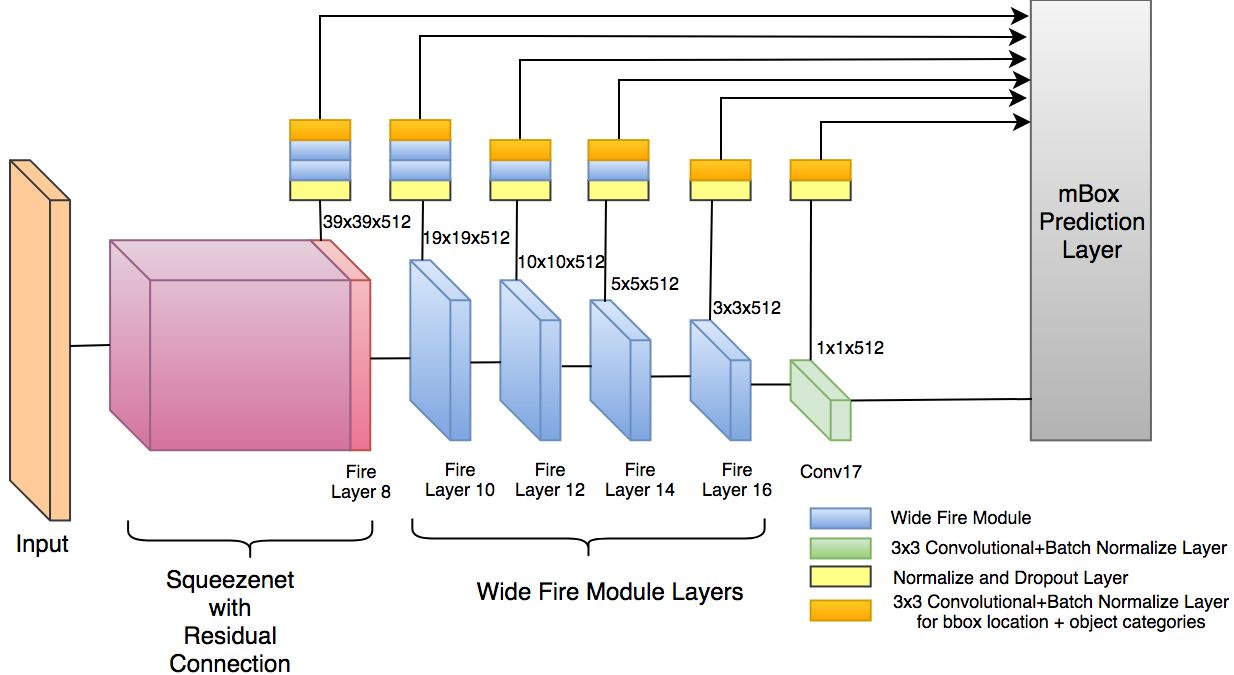}
\caption{Network architecture of Fire SSD}
\label{figurelabel1}
\end{center}
\vspace{-2mm}
\end{figure*}

\subsection{Wide Fire Module}

WFM shown in Figure3(a) is inspired by SqueezeNet, ResNext and ShuffleNet. group convolution not only reduces the computational complexity, but also improves the accuracy at the same time.  We improved the fire module by replacing the $3\times3$ and $1\times1$ expand convolution with group convolution. The study conducted in ResNext shows that when the network is going wider and increasing the cardinality, it has a better results than going deeper or wider in images classification task.  Compared to going wider and deeper, increasing cardinality only comes with minimal increase in terms of computational complexity. One can view this as ensembling multipath of $3\times3$ convolution and $1\times1$ convolution filters. 
For Fire SSD, the cardinality of $3\times3$ group convolution is set to 16 and the cardinality of $1\times1$ group convolution is set to 2. Higher cardinality for $1\times1$ group convolution has been tested and it shows that the performance improvement is minimal and the performance is decreased. We argues that this is due to the receptive field of $1\times1$ convolution which is much smaller than $3\times3$ convolution. Further breaking the $1\times1$ convolution into multiple groups of smaller number of channel features may not be able to capture the useful information from the input feature maps.  To balance the number of features generated by group  $1\times1$ and $3\times3$ convolution, the cardinality is set based to fulfill the conditions in (1).
$$
C_{1x1}\times K_{1x1} \approx C_{3x3}\times K_{3x3}  \eqno{(1)}
$$

where $C$ is the number of channels for each grouped convolution and $K$ is the filter size.

\subsection{Dynamic Mbox Detection Layer}

In SSD framework, six feature maps with different resolutions are extracted from the backend network. Each feature map is processed by a single $3\times3$ convolutional layer for bounding box regression and object class prediction.  Feature maps at higher layers are responsible for detecting small scale objects. SSD does not perform well on detecting small scale objects because the corresponding feature maps are not deep enough and extracted features are not semantic enough for detecting objects. VGG16 and ResNet show that the deeper features can improve the classification accuracy.  

To overcome this problem, we proposed dynamic residual mbox detection Layer(DRMD). We stacked different number of convolutional layers for feature maps based on the feature map size to increase the depth of each mbox branches.   To keep the model size and computation complexity small, WFM is used to replace conventional $3\times3$ convolution layer. To compensate the gradient vanishing effect as the number of convolutional layers grow, residual connection is introduced to connect the upper and lower WFM modules. The residual connection only applied to $38\times38$ and $19\times19$ feature maps because the element wise sum layer requires heavy memory operation. The details of dynamic residual mbox detection layer are presented in Figure 4. For $38\times38$ and $19\times19$ feature maps, two WFMs are introduced before the $3\times3$ convolution layer.    For $10\times10$ and $5\times5$ feature maps, one WFM is introduced before the $3\times3$ convolution layer. The stem block with different number of WFM modules simultaneously learn the features that are required to predict bounding box location and object class. 2 $3\times3$ convolution layers are appended at end of the stem block. One for bounding box prediction and another one for object class classification. No WFM is added to $3\times3$ and $1\times1$ feature map branch as large objects are easier to be detected.

\subsection{Normalize and Dropout module}

In SSD300, the objective functions are applied directly on the selected feature maps and a L2 normalization layer is used for the conv4\_3 layer. It is due to the large magnitude of the gradient. We investigated the layers that contain mbox branch and found that the gradient magnitude is different for each layer. Keeping this in mind, we added batch normalize layer for each subnet. The filter size for each layer is designed in such a way that it is fixed to 512 channels to ensure that mbox layers obtain sufficient information to make decision. Large number of filters come with high risk of overfitting. To avoid overfitting while using such large number of filters, appended dropout layer is implemented after batch normalize layer. The dropout ratio is set to a small value to minimize the variance shift problem~\cite{AdjustDropout}. Dropout layers in each subnet regularize the network by adding noise to the input feature map for each mbox subnet. One can view this as data augmentation on training data. The dropout ratio varies according to the size of the feature map. Larger feature maps have larger dropout ratio because the larger the number of parameters the more it is prone to  variance shift. We named the sub-network design a normalized and dropout module (NDM) as shown in Figure 3(b).  

\begin{figure}[ht]
\centering
\includegraphics[scale=0.5]{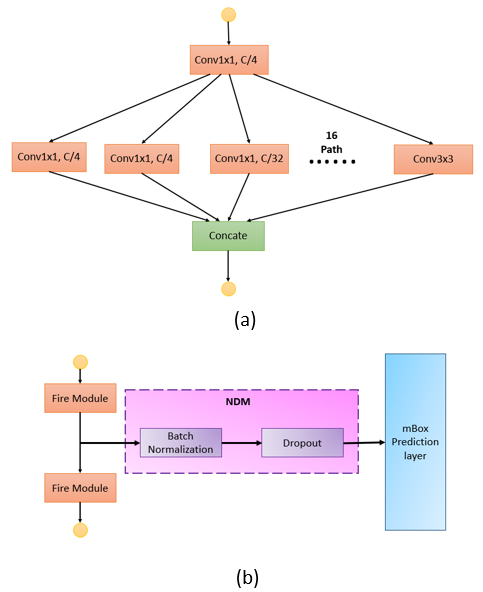}
\caption{(a) Wide Fire Module. (b) Normalization and Dropout Module}
\label{figurelabel}
\end{figure}

\section{Experiment}

The experiments are conducted on the widely used PASCAL VOC dataset that has 20 object categories.  Object detection accuracy is measured by mean Average Precision (mAP).

\subsection{Network Training}

The model is trained on the union of PASCAL 2007 trainval and 2012 trainval datasets with Caffe~\cite{caffe}.  The backend network of Fire SSD is SqueezeNet with residual connection. The newly added layers are initialized with Xavier method~\cite{UDTDFNN}. The training process consists of four stages. The optimization algorithm for training is stochastic gradient decent (SGD).  During the first stage of training, the backend network is frozen from updating. The learning rate is warmup~\cite{SGD} to 0.001 and batch size is 64 for the first 50K iterations. For the second stage, the backend network is unfrozen. The learning rate is adjusted to 0.01 and batch size is increased to 128 for another 50K iterations. During the third stage, the batch size is increased to 256 and learning rate remains the same for another 25k iterations. During the fourth stage of training, the learning rate is reduced to 0.001 for another 30k iterations to fine-tune the result.  

\subsection{Effect of Various Design Choice}

Table 2 summarize the ablation results on different sub-modules. By introducing WFM to a new layers, it reduced the parameters and MACs by 9.3\% and 6\% respectively while improving the accuracy by 0.4mAP. Adding the DRMD, the number of parameters and MACs increases but it  improves the performance of detecting small scale objects significantly as shown in Figure 1.  Finally, with the help of NDM to regularized the network training, the performance boosted by 1.4mAP.  In terms MACs, Fire SSD should be about 2.3 times slower than SSD+MobileNet but in practical the speed is 2-4 times slower. This is due to lack of efficiency implementation of group convolution. 

\begin{table}[ht]
\caption{Effect of various design choices}
\begin{center}
\begin{tabular}{ |p{2.2cm}|p{1cm}|p{1cm}|p{1cm}|p{1cm}|}
\hline
Dynamic Residual Mbox Detection Layer & $\times$ & $\times$  & - & -  \\
\hline
Wide Fire Module & $\times$  & $\times$  & $\times$  & - \\
\hline
Normalize Dropout Module & $\times$  & - & - & - \\
\hline
\hline
\textbf{Number of Parameters} & 7.13M & 7.13M & 6.77M & 7.94M\\
\hline
\textbf{MACs} & 2.67G & 2.67G & 2.43G & 2.59G\\
\hline
\textbf{VOC 2007 mAP} & \textbf{70.5} & 69.1 & 68.9 & 68.5\\
\hline
\end{tabular}
\end{center}
\end{table}

\subsection{Pascal VOC 2007}

As can be seen from the Table 3, among the lightweight object detectors, Fire SSD scores the highest accuracy. It has higher accuracy  than YOLOv2 by 2.7mAP with only about 10\% of YOLOv2's model size. In terms of model size, Fire SSD is about 2 times smaller than  Tiny YOLOV2's model size but achieves 13.4\% higher accuracy. Compared with the other SSD like implementations, Fire SSD outperforms SSD+MobileNet by 2.5mAP and SSD+SqueezeNet by 6.2mAP. The comparisons are shown in Table 3.

\begin{table}[ht]
\caption{PASCAL VOC 2007 Test Result}
\begin{center}
\begin{tabular}{ |p{1.8cm}|p{2cm}|p{2cm}|p{1cm}|}
\hline
\textbf{Model} & \textbf{Computation Cost (Millions MACS)} & \textbf{Model Size (Million Parameters)} & \textbf{Accuracy (mAP)}\\
\hline
YOLO V2 & 8,360 & 67.13 & 69\\
\hline
Tiny YOLO V2 & 3,490 & 15.86 & 57.1\\
\hline
SSD + MobileNet~\cite{SATradeOff} & \textbf{1,150} & 5.77 & 68\\
\hline
SSD + SqueezeNet~\cite{SATradeOff} & 1,180 & \textbf{5.53} & 64.3\\
\hline
\textbf{Fire SSD} & 2,670 & 7.13 & \textbf{70.5}\\
\hline
SSD300 & 31,380 & 26.28 & 77.2\\
\hline
\end{tabular}
\end{center}
\end{table}

\subsection{Inference Speed on CPU and GPU}

We tested Fire SSD on a small form factor computer called Intel\textsuperscript{\tiny\textregistered} Next Unit of Computing (NUC). The model selected is Skull Canyon~\cite{Skull}.  Skull Canyon features a quad core CPU and integrated GPU.  The size of Skull Canyon is $211mm \times 116mm \times 28mm$ which is ideal for edge deployment. Besides that, we also run the network on  Movidous\textsuperscript{\tiny\textregistered} Neural Compute Stick (NCS)~\cite{myriad}.

\begin{itemize}
\item \textbf{CPU}. Skull Canyon features a sixth generation Core i7 processor. The CPU running at 45W and clocked at 2.6GHz. The CPU consists of 4 cores and support AVX2 instruction set which can speed out the network inference time. The low power CPU can be deployed in small form factor and yet powerful enough to handle object detection work load.

\item \textbf{GPU}. The GPU of Skull Canyon is Intel\textsuperscript{\tiny\textregistered} Iris\texttrademark Pro Graphics 580. The GPU consists of 72 execution units and clocked at 350MHz. The GPU can be programmed by OpenCL\texttrademark API.

\item \textbf{VPU}. NCS  is a tiny fanless deep learning device that can be use to accelerate computer vision work load. The VPU of NCS consists of 12 VLIW 128-bit vector processors optimized for machine vision. NCS only drawing 1W power which makes it ideal for very low power edge devices.
\end{itemize}

Fire SSD is implemented in IntelCaffe~\cite{CaffeOpt} and OpenVINO\texttrademark~\cite{OpenVINO}. IntelCaffe is a fork from Caffe that is optimized for Intel\textsuperscript{\tiny\textregistered} CPU. OpenVINO\texttrademark supports deep learning acceleration on different Intel hardware platforms. OpenVINO\texttrademark also optimizes the the network model using well-know optimization techniques such as fusing convolutional and activation layers, fusing group convolution and reducing the floating point's precision of network parameters.

In this paper, we evaluated the inference throughput of network on CPU, integrated GPU and VPU. The results are presented in Table 4. The batch size for all tests are set to  1. Fire SSD achieves 11.5FPS using IntelCaffe.  Using OpenVINO\texttrademark, Fire SSD achieves 31.7FPS and meet the real-time requirement. The improved speed gained is due to build-in automatic optimizer of OpenVINO\texttrademark.  For GPU, OpenVINO\texttrademark supports both FP32 and FP16 computations. Fire SSD scores 33.1FPS running on GPU. Reducing the floating point precision to FP16, the Fire SSD's GPU score reaches 39.8FPS. Fire SSD also scores 3FPS on VPU. Although Fire SSD is not the fastest network, it is the only network  that scores beyond 70mAP and achieves real-time object detection running on off-the-shelf CPU. 

\begin{table}[ht]
\caption{Inference Speed on CPU and GPU}
\begin{center}
\begin{tabular}{ |p{1.2cm}|p{0.9cm}|p{1.1cm}|p{0.6cm}|p{0.6cm}|p{0.7cm}|p{0.6cm}|}
\hline
&&&\multicolumn{4}{c|}{OpenVINO\texttrademark} \\ \cline{4-7}
\textbf{Model} & \textbf{Accuracy (mAP)} & \textbf{IntelCaffe CPU} & \textbf{CPU} & \textbf{GPU} & \textbf{GPU (FP16)} & \textbf{VPU}\\
\hline
Tiny YOLO V2 & 57.1 & NA & 49.5 & \textbf{84.2} & \textbf{108.6}  & 6.6\\
\hline
SSD + MobileNet & 68 & 27.8 & \textbf{91.7} & 69.4 & 90.9  & 12.5\\
\hline
SSD + SqueezeNet & 64.3 & \textbf{35.7} & 80 & 57.8 & 67.6  & 7.1\\
\hline
\textbf{Fire SSD} & \textbf{70.5} & 11.5 & 31.7 & 33.1 & 39.8 & 2.9\\
\hline
\hline
SSD300 & 77.2 & 4.2 & 5.2 & 10.8 & 18 & 0.7 \\
\hline
\end{tabular}
\end{center}
\end{table}

\section{CONCLUSIONS}

In this work, we present three new network modules to enhance the accuracy of SqueezeNet based SSD. The  three new models augment the detector in different aspects. Our model achieves 70.5mAP accuracy on PASCAL VOC 2007 while keeping real-time speed performance on CPU and integrated GPU. Our model hits the sweet spot for the between the CPU speed and accuracy. Besides that, the model size  of Fire SSD is just 28MB and can be distributed easily over network. Our model will be useful to edge devices especially edge device with CPU only.

\addtolength{\textheight}{-12cm}   % This command serves to balance the column lengths

%%%%%%%%%%%%%%%%%%%%%%%%%%%%%%%%%%%%%%%%%%%%%%%%%%%%%%%%%%%%%%%%%%%%%%%%%%%%%%%%
%\bibliographystyle{}
\bibliographystyle{unsrt}
\bibliography{references}
\end{document}